\documentclass[sigconf]{acmart}
\AtBeginDocument{%
  }

\setcopyright{acmlicensed}
\copyrightyear{2026}
\acmYear{2026}
\setcopyright{cc}
\setcctype{by-nc-nd}
\acmConference[SIGIR '26]{Proceedings of the 49th International ACM SIGIR Conference on Research and Development in Information Retrieval}{July 20--24, 2026}{Melbourne, VIC, Australia}
\acmBooktitle{Proceedings of the 49th International ACM SIGIR Conference on Research and Development in Information Retrieval (SIGIR '26), July 20--24, 2026, Melbourne, VIC, Australia}
\acmDOI{10.1145/3805712.3809637}
\acmISBN{979-8-4007-2599-9/2026/07}

\usepackage[utf8]{inputenc}
\usepackage{array}
\usepackage{CJKutf8}
\usepackage{hyperref}
\usepackage{microtype}
\usepackage{tcolorbox}
\usepackage{multirow}
\definecolor{LimeGreen}{rgb}{0.2, 0.8, 0.2}
\definecolor{Red}{rgb}{1, 0, 0}

\begin{document}

\title{Decoding Multimodal Cues: Unveiling the Implicit Meaning Behind Hateful Videos}

\author{Junyu Lu}
\authornote{Junyu Lu completed this work while interning at Tencent as part of the Tencent Rhino-Bird Research Elite Program, with Deyi Ji as the program leader.}
\affiliation{%
  \institution{Dalian University of Technology}
  \city{Dalian}
  \country{China}}
\email{dut_ljy@foxmail.com}

\author{Deyi Ji}
\authornote{Corresponding author.}
\affiliation{%
  \institution{Tencent}
  \city{Beijing}
  \country{China}}
\email{deyiji@tencent.com}

\author{Liqun Liu}
\affiliation{%
  \institution{Tencent}
  \city{Beijing}
  \country{China}}
\email{liqunliu@tencent.com}

\author{Xiaokun Zhang}
\affiliation{%
  \institution{City University of Hong Kong}
  \city{Hong Kong}
  \country{China}}
\email{dawnkun1993@gmail.com}

\author{Youlin Wu}
\affiliation{%
  \institution{Dalian University of Technology}
  \city{Dalian}
  \country{China}}
\email{mzlwuyoulin@gmail.com}

\author{Roy Ka-Wei Lee}
\affiliation{%
  \institution{Singapore University of Technology and Design}
  \city{Singapore}
  \country{Singapore}}
\email{s.roylee@gmail.com}

\author{Peng Shu}
\affiliation{%
  \institution{Tencent}
  \city{Beijing}
  \country{China}}
\email{archershu@tencent.com}

\author{Huan Yu}
\affiliation{%
  \institution{Tencent}
  \city{Beijing}
  \country{China}}
\email{huan.yu@gmail.com}

\author{Jie Jiang}
\affiliation{%
  \institution{Tencent}
  \city{Beijing}
  \country{China}}
\email{zeus@tencent.com}

\author{Bo Xu}
\affiliation{%
  \institution{Dalian University of Technology}
  \city{Dalian}
  \country{China}}
\email{xubo@dlut.edu.cn}

\author{Liang Yang}
\affiliation{%
  \institution{Dalian University of Technology}
  \city{Dalian}
  \country{China}}
\email{liang@dlut.edu.cn}

\author{Hongfei Lin}
\authornotemark[2]
\affiliation{%
  \institution{Dalian University of Technology}
  \city{Dalian}
  \country{China}}
\email{hflin@dlut.edu.cn}

\renewcommand{\shortauthors}{Junyu Lu et al.}

\begin{abstract}
Hateful videos have become prevalent on online platforms, highlighting an urgent need for effective detection. However, existing studies primarily focus on binary classification and fail to provide contextual rationales that reveal the implicit meanings behind these judgments, significantly undermining model explainability. To fill this gap, we aim to achieve explainable hateful video detection, enabling models to provide contextual rationales that integrate relevant evidence and logical reasoning alongside decisions. This approach can comprehensively enhance the understanding of video content and the explainability of the decision-making process. We first introduce two datasets, Ex-HateMM and Ex-ImpliHateVid, for explainable hateful video detection. Each dataset provides fine-grained annotations of multimodal harmful elements, along with contextual rationales. We then propose an \textbf{I}nformation \textbf{A}ugmentation and \textbf{R}easoning \textbf{E}nhancement (\textbf{IARE}) framework designed for explainable detection. The framework employs an \textbf{information augmentation phase} that leverages the multimodal chain-of-thought to integrate harmful elements, thereby enriching rationale evidence. Additionally, IARE incorporates a \textbf{reasoning enhancement phase}, in which Direct Preference Optimization guides the model toward correct reasoning paths and away from incorrect ones, thereby improving the logical coherence of its justifications. We conduct extensive experiments on the two datasets, comparing multiple baselines with our proposed IARE framework. The results demonstrate that IARE achieves state-of-the-art performance while also generating accurate rationales.

% employs information entropy regularization and gated fusion to balance
% This knowledge is then utilized to fine-tune a small language model as the detector.
% Furthermore, a hybrid expert network integrates these two types of information to ensure a comprehensive understanding of memes.
% This knowledge is then incorporated into a small language model for final detection decisions.
% , introducing precise knowledge of memes and fully integrating intrinsic information for decision-making
% Extensive experiments are conducted on three publicly available datasets of hateful memes. 
% The results demonstrate that M2KE significantly enhances meme comprehension, achieving state-of-the-art performance across all datasets. 
% Further analysis underscores the significance of accurately introducing and integrating knowledge to improve model performance and explainability.

\end{abstract}

\begin{CCSXML}
<ccs2012>
 <concept>
  <concept_id>00000000.0000000.0000000</concept_id>
  <concept_desc>Do Not Use This Code, Generate the Correct Terms for Your Paper</concept_desc>
  <concept_significance>500</concept_significance>
 </concept>
 <concept>
  <concept_id>00000000.00000000.00000000</concept_id>
  <concept_desc>Do Not Use This Code, Generate the Correct Terms for Your Paper</concept_desc>
  <concept_significance>300</concept_significance>
 </concept>
 <concept>
  <concept_id>00000000.00000000.00000000</concept_id>
  <concept_desc>Do Not Use This Code, Generate the Correct Terms for Your Paper</concept_desc>
  <concept_significance>100</concept_significance>
 </concept>
 <concept>
  <concept_id>00000000.00000000.00000000</concept_id>
  <concept_desc>Do Not Use This Code, Generate the Correct Terms for Your Paper</concept_desc>
  <concept_significance>100</concept_significance>
 </concept>
</ccs2012>
\end{CCSXML}

\ccsdesc[500]{Computing methodologies~Artificial intelligence; Natural language processing}

\keywords{Hateful video detection, Large language model, Multimodal understanding, Reasoning Enhancement}


\maketitle

\section{Introduction}

\begin{figure*}[htpb]
    \centering
    \includegraphics[width=17.5cm]{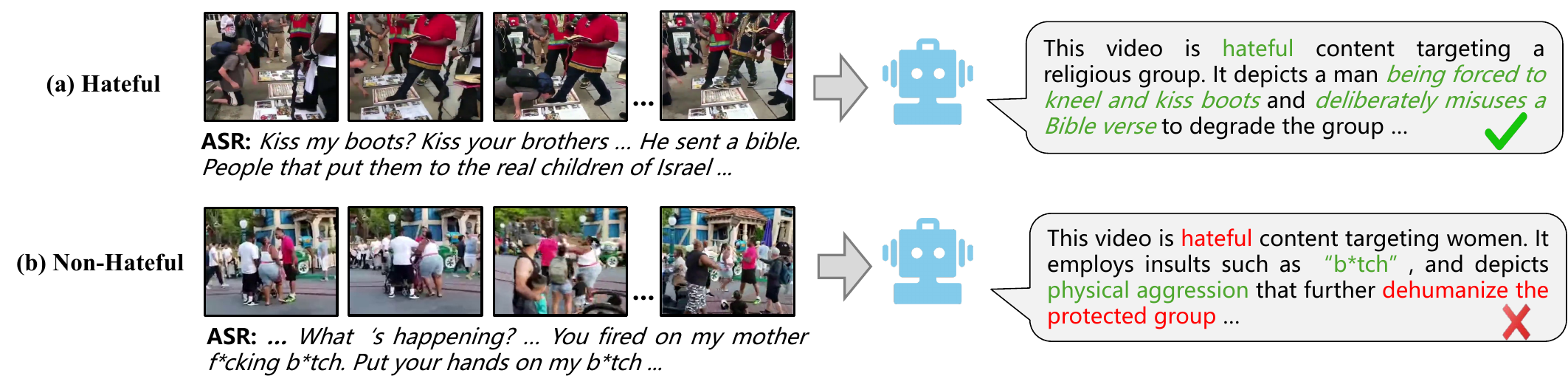}
    \vspace{-0.1in}
    \caption{Illustration of explainable hateful video detection, where contextual rationales are provided alongside predictions. Accurate predictions and supporting cues are highlighted in \textcolor{LimeGreen}{green}, while inaccuracies or spurious evidence are marked in \textcolor{Red}{red}.}
    \vspace{-0.1in}
    \label{intro}
\end{figure*}

With the rapid advancement of the internet, videos have become a powerful medium for emotional expression and public communication \cite{DBLP:conf/sigir/VannesteVG24}. 
While they bring considerable convenience and enrichment to people’s lives, the proliferation of hateful videos has also become increasingly prominent across online platforms.
Hateful videos\footnote{\textcolor{red}{\textbf{\textit{Disclaimer}}: The samples presented by this paper may be considered profane, offensive, or hateful.}} are typically defined as multimedia content that attacks, demeans, or incites hostility toward individuals or groups based on protected attributes such as race, religion, gender, or physical characteristics \cite{DBLP:conf/icwsm/DasRSMG023}.
Such content often reinforces harmful stereotypes, leading to discrimination, harassment, or even physical violence against targeted groups, with profound negative impacts on both individuals and society.
Therefore, it is imperative to develop automated methods to effectively detect hateful video content.

In recent years, substantial progress has been achieved in the field of hateful video detection, driven by the integration of multimodal and natural language processing techniques. 
This advancement has been supported by the introduction of benchmark datasets, such as HateMM \cite{DBLP:conf/icwsm/DasRSMG023} and ImpliHateVid \cite{DBLP:conf/acl/RehmanBKBK25}, which in turn have facilitated the development of various detection methods \cite{DBLP:conf/www/LangHXLX025, DBLP:conf/www/KoushikKT25, DBLP:conf/www/WangTL25}.
However, most existing studies remain limited to binary classification (i.e., predicting whether a video is hateful or not) and fail to provide corresponding decision rationales, resulting in limited model explainability.
This significantly hinders the practical deployment of such models in real-world applications, since content moderators and users often require understandable justifications for why a sample is flagged as hateful \cite{DBLP:conf/ijcai/HeeCL23, DBLP:conf/www/LinLGMWY24}.

In this paper, we focus on the development of explainable hateful video detection.
We aim to achieve accurate prediction performance while generating contextual rationales for the model’s decisions.
During explainable detection, these rationales are intended to improve both the understanding of video content and the explainability of the model’s decisions.
To this end, we consider the following two principles to design our model:
First, the rationales must be grounded in sufficient evidence, including specific multimodal elements drawn from the video.
For instance, in Figure~\ref{intro}(a), a persuasive rationale should integrate the insulting phrase “\textit{Kiss my boots}” with the religious scripture to clearly explain the hateful intent conveyed in the video.
Second, rationales should be organized based on logical reasoning rather than overly relying on certain elements.
For example, in Figure~\ref{intro}(b), if the model oversimplifies the association between insults and violence as inherently implying hateful intent, it may misclassify this sample as hate speech, even though it merely depicts a non-targeted physical fight rather than hatred toward a particular group.

To enhance detection accuracy and ensure high-quality rationales, our efforts focus on two aspects: dataset construction and detector development. 
We first build two datasets, Ex-HateMM and Ex-ImpliHateVid, by extending existing open-sourced hateful video datasets \cite{DBLP:conf/icwsm/DasRSMG023, DBLP:conf/acl/RehmanBKBK25}. 
In extending the dataset, we emphasize the separation of objective facts and subjective inferences to ensure data reliability. 
Specifically, we perform fine-grained annotation, including video caption, potentially harmful elements across different modalities, such as offensive language and unsafe scenes in the video, and the underlying contextual rationale that analyzes why the video is hateful or non-hateful.
To the best of our knowledge, these are the first resources tailored to explainable hateful video detection, which provide a data foundation for subsequent model training and evaluation.

Furthermore, we propose a novel \textbf{I}nformation \textbf{A}ugmentation and \textbf{R}easoning \textbf{E}nhancement (\textbf{IARE}) framework for explainable hateful video detection.
First, we design an \textbf{information augmentation stage} to enrich the generated rationale.
This stage employs a multimodal chain-of-thought mechanism, guiding reasoning by integrating rich contextual cues, including video captions and identified potentially harmful elements.
This ensures that the output rationales are firmly grounded in specific evidence, while also deepening the model's understanding of videos through these elements.
Next, we propose a \textbf{reasoning enhancement stage} to further refine the model's ability to distinguish between hateful and non-hateful content.
In this stage, the model is presented with both correct reasoning paths and erroneous ones that spuriously link contextual cues to incorrect labels, and is required to make the final decision. 
We train the model with Direct Preference Optimization (DPO), encouraging it to prefer logically sound rationales over spurious associations, thereby improving the coherence and reliability of its explanations. 
Through this two-stage optimization, our proposed IARE framework improves both detection accuracy and the explainability of model decisions.

The contributions of this paper are summarized as follows:
\begin{itemize}
\item We construct two explainable hateful video datasets, Ex-HateMM and Ex-ImpliHateVid. To the best of our knowledge, they are the first datasets for explainable detection. We provide fine-grained annotations that identify harmful elements across modalities and capture the underlying rationales supporting hateful and non-hateful judgments\footnote{The datasets and codes are available at \url{https://github.com/DUT-lujunyu/IARE}}.
\item We propose an Information Augmentation and Reasoning Enhancement (IARE) framework for explainable hateful video detection. It consists of an information augmentation phase to incorporate rich contextual information of video content, and a reasoning enhancement phase to improve the logical coherence of the justifications.
\item We utilize the two datasets to evaluate the detection performance and explanation generation capabilities of multiple baseline models and our proposed IARE. 
The results demonstrate that IARE achieves state-of-the-art (SOTA) performance.
Detailed fine-grained analysis further illustrates the explainability of IARE.
% Furthermore, we summarize the challenges in explainable hateful video detection.
\end{itemize}

\section{Related Work}

\subsection{Explainable Hate Speech Detection}
Researchers have made substantial progress in hate speech detection, driven by its severe harms to both individuals and society.
With continued research progress, the explainability of detection model decisions has received increasing attention, with rationale generation emerging as a central approach for assessing model reliability \cite{DBLP:journals/jocss/GonganeMA24}.
\cite{DBLP:conf/emnlp/ElSheriefZMASCY21, DBLP:conf/acl/SapGQJSC20} and \cite{DBLP:conf/ijcai/HeeCL23} have constructed explainable hate speech and hateful meme datasets, respectively, providing underlying rationales behind hateful content.  
Additionally, various reasoning techniques, including chain-of-thought \cite{DBLP:conf/emnlp/YangKKHTY23, lu2024towards}, debate \cite{DBLP:conf/www/LinLGMWY24}, and multi-agent collaboration \cite{wang2026multi, DBLP:conf/sigir/Lu00ZW0L25}, are employed to improve the quality of rationales generated by LLMs for hateful examples, thereby enhancing the explainability of model predictions.
Nevertheless, explainable hateful video detection remains largely underexplored. Unlike text or memes, videos convey meaning through temporal dynamics and richer contextual cues \cite{10982110}, which makes explainable detection substantially more challenging. In this paper, we aim to address this critical research gap.

\begin{table*}
  \centering
  \caption{Statistics of Ex-HateMM and Ex-ImpliHateVid dataset, including the dataset size, video length, the average text length of each field (video text, caption, rationale), and the number of harmful elements.}
  % \vspace{-0.025in}
    \begin{tabular}{c|ccc|ccccc}
    \toprule
    Dataset & Hate  & Non-Hate & Total & Avg. video len. & Avg. text len. & Avg. caption len. & Avg. elem. & Avg. rationale len. \\
    \midrule
    Ex-HateMM & 419   & 651   & 1,070  & 144.90 & 283.27 & 40.04 & 3.98  & 72.68 \\
    Ex-ImpliHateVid & 1,007  & 998   & 2,005  & 142.11 & 107.38 & 39.90  & 2.37  & 60.90 \\
    \bottomrule
    \end{tabular}%
  \label{statistics}%
\end{table*}%

\subsection{Hateful Video Detection}
Current research in multimodal hateful video detection largely focuses on optimizing cross-modal feature representation and fusion mechanisms \cite{ji2025raven, ji2025raven++}.
Early studies directly employed existing unimodal and multimodal backbone architectures for detection \cite{DBLP:conf/lrec/AlcantaraMF20, DBLP:conf/icwsm/DasRSMG023, DBLP:conf/mm/WangYNL24, Wang2024TowardsPA}.
Subsequent work has further explored enhanced cross-modal interaction mechanisms to improve detection performance \cite{DBLP:conf/www/LangHXLX025, DBLP:conf/www/KoushikKT25, cespedes2025mm, DBLP:conf/www/LangHXLX025}.
Recently, \cite{DBLP:conf/www/WangTL25} proposed a transfer learning framework that leverages existing hateful meme datasets to pre-train models, thereby enhancing their generalization capability in hateful video detection tasks.
\cite{DBLP:conf/acl/RehmanBKBK25} focused on identifying implicitly hateful videos, which are more subtle and semantically complex, by incorporating auxiliary information, including video captions and sentiment cues, to improve the model's comprehension of videos.
Despite this progress, the black-box nature of these models still constrains decision explainability. 
To address this, we require models to provide corresponding rationales alongside their decisions, thereby enhancing transparency and promoting explainability.

\subsection{Large Language Model}
In recent years, large language models (LLMs) have rapidly advanced, demonstrating extensive world knowledge and strong generative capabilities \cite{DBLP:conf/nips/KojimaGRMI22, DBLP:journals/jmlr/ChowdheryNDBMRBCSGSSTMRBTSPRDHPBAI23, DBLP:journals/corr/abs-2303-08774, DBLP:journals/corr/abs-2201-08239}. Early studies indicate that LLMs have become a key backbone for the research of natural language processing, substantially improving downstream models’ ability to understand and represent textual information \cite{DBLP:conf/iclr/0002WSLCNCZ23}. 
In parallel, dedicated efforts have sought to further enhance their reasoning capabilities \cite{DBLP:journals/corr/abs-2309-15402, DBLP:conf/iclr/0002WSLCNCZ23}. 
As research has progressed, increasing work has extended LLM reasoning to multimodal tasks, such as visual question answering \cite{DBLP:journals/csur/KuangSXLXLLCLH25} and multimodal understanding \cite{10982110, DBLP:conf/sigir/Lu00ZW0L25, 10.1145/3774904.3793046}.
Additionally, some studies have applied LLMs to dataset construction by adopting a human-in-the-loop paradigm to assist manual annotation, substantially reducing labor costs and improving efficiency \cite{DBLP:conf/chi/Wang0RMM24, DBLP:conf/emnlp/TanLWBJBKL0024, DBLP:conf/eacl/KimMCRZ24}. In this paper, we leverage LLMs to facilitate the construction of the hateful video datasets and further propose an explainable framework for hateful video detection.

\section{Dataset Construction}


\subsection{Overall}
\textbf{Problem Statement}: In this paper, we reformulate the binary classification task of hateful video detection as a generative problem, leveraging the generative capabilities of LLMs.
We refer to existing studies on hateful video detection \cite{DBLP:conf/acl/RehmanBKBK25, DBLP:conf/www/LangHXLX025} that convert both audio information and textual elements from video frames into text, thereby bridging the semantic gap between different modalities. 
As a result, each sample in the hateful video dataset is a tuple comprising textual ($T$) and visual ($I$) modalities, which serve as the input to the model. 
The model generates a textual output label $y\in$ \{hate, non-hate\}, indicating whether the video is hateful, along with contextual rationales $r$ supporting the decision.

In the following section, we provide a detailed description of the construction process for the explainable hateful video datasets, Ex-HateMM and Ex-ImpliHateVid. 
We first introduce the source datasets, followed by our data preprocessing pipeline. We then present the fine-grained annotation procedure, which provides contextual information and evidence-grounded rationales.

\subsection{Source Datasets}
Due to ethical and copyright considerations, we refrain from directly scraping or collecting data from online platforms.
Instead, we expand the annotations of two established, peer-reviewed hateful video datasets: HateMM \cite{DBLP:conf/icwsm/DasRSMG023} and ImpliHateVid \cite{DBLP:conf/acl/RehmanBKBK25}. 
These datasets are primarily sourced from two English-language social video-sharing platforms, BitChute and Odysee.
In contrast to mainstream platforms such as YouTube, these platforms promote diverse, decentralized content and apply relatively limited moderation. 
As a result, they host many creators who have been banned from more strictly moderated services and exhibit a higher prevalence of hateful material.
Additionally, although other datasets exist \cite{DBLP:conf/lrec/AlcantaraMF20, DBLP:conf/www/WangTL25}, we exclude them because they provide only textual excerpts or links to video files, rather than directly usable multimodal content, which prevents straightforward re-annotation.

\subsection{Data Preprocessing}

To facilitate subsequent research, we performed a systematic preprocessing procedure on the video files in the original datasets. 
First, we conducted a quality check on the video data, using the ffmpeg tool for decoding verification to exclude corrupted files and ensure data integrity. 
On this basis, we performed equidistant temporal sampling on the videos to construct keyframe sequences. 
Subsequently, we applied Whisper-large\footnote{http://github.com/openai/whisper} for automatic speech recognition (ASR) and PaddleOCR\footnote{https://github.com/PaddlePaddle/PaddleOCR} for optical character recognition (OCR) to extract textual information from the audio track and video frames, respectively. Both tools are widely used in video processing. We further incorporated manual verification to ensure the reliability of the extracted text. 
On the two datasets, the ASR and OCR accuracies, evaluated on 500 randomly sampled instances, are 96.5\% and 94.0\% on average, respectively, suggesting that the extracted textual content is reliable for subsequent analysis.

\subsection{Data Annotation}

During annotation, we explicitly separate objective facts from subjective inferences, ensuring data reliability. Specifically, we provide fine-grained labels including (i) video captions, (ii) potentially harmful elements across modalities, and (iii) contextual rationales that justify why a video is classified as hateful or non-hateful. The annotation guidelines are summarized as follows:

\textbf{Video Caption}: Annotators are first required to objectively and neutrally summarize the main video content, describing observable scenes and actions. To minimize the influence of personal bias, any subjective interpretations or inferences must be avoided in this annotation stage.

\textbf{Potential Harmful Elements}: Then, specific elements that present potential safety risks are identified from the video. We respectively analyze the textual and visual information in the video. For textual information, we extract offensive or sensitive vocabulary by referencing an existing lexicon of insulting terms. For visual information, four major categories of harmful content are focused on, including violence (e.g., weapons), hate (e.g., Nazi symbols), pornography (e.g., sexual acts), and illegal activities (e.g., drugs), as defined by the existing security framework \cite{DBLP:journals/corr/abs-2503-17682, DBLP:journals/ijcv/YingLLHGZLT26}.

\textbf{Contextual Rationales}: Finally, annotators provide an explanation supporting the binary classification decision, grounded in the video content and relevant contextual information, including the annotated potentially harmful elements. For hateful videos, annotators must explicitly identify the targeted group and provide a concise justification based on the corresponding harmful elements. For non-hateful videos, annotators should summarize the video’s primary theme or intent.

To support annotators in video analysis and text refinement, we employ GPT-4o-mini as an auxiliary tool in the implementation, improving annotation throughput while maintaining high standards of clarity and consistency. To further ensure annotation quality, we establish a rigorous quality-control process. Each sample is independently annotated by two senior annotators (Master’s-level), and their outputs are then verified and consolidated by a junior expert (Ph.D.-level), who determines the final label.
During quality verification, since these annotation tasks involve generative or extractive labeling rather than simple classification, traditional agreement metrics (e.g., Cohen’s $\kappa$) are not directly applicable. 
We therefore require the junior expert to assess the consistency between the two senior annotators using a five-point Likert scale. Across three dimensions, i.e., \textit{Video Caption}, \textit{Potentially Harmful Elements}, and \textit{Contextual Rationales}, the average consistency scores are 4.75, 4.13, and 4.46, respectively, reflecting strong overall agreement in the annotators’ understanding of each sample.

After consolidating the annotations at all granularities, we construct the Ex-HateMM and Ex-ImpliHateVid. The final statistics of the two datasets are reported in Table~\ref{statistics}.

\begin{figure*}[htpb]
    \centering
    \includegraphics[width=17.75cm]{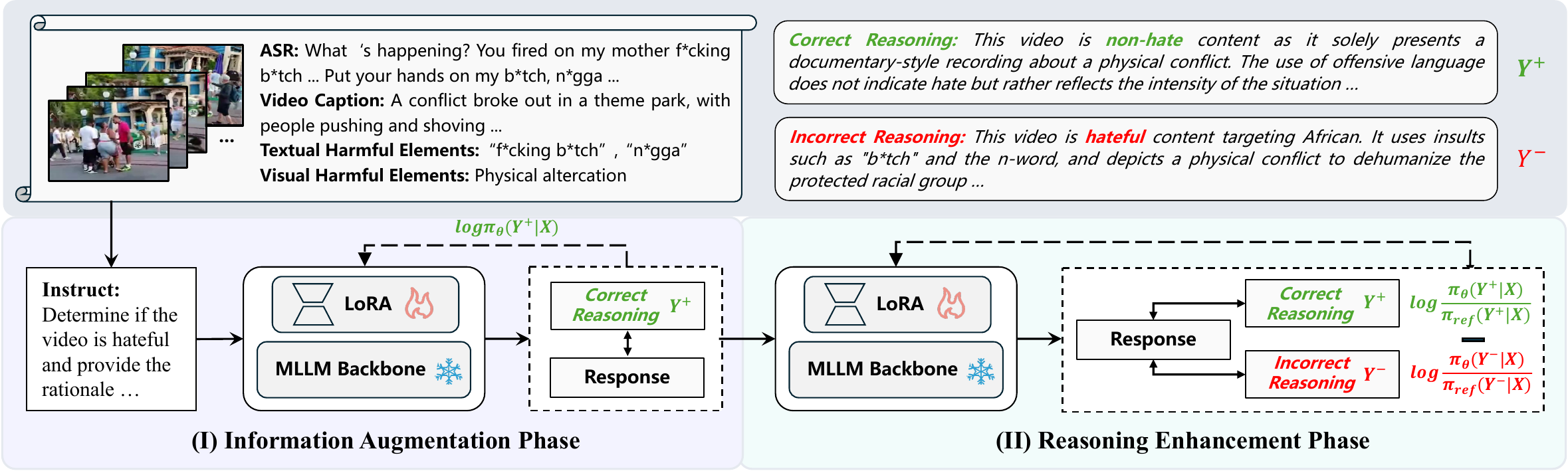}
    % \vspace{-0.025in}
    \caption{Overall illustration of our proposed IARE framework. We first design an information augmentation phase to incorporate rich contextual information of video content, followed by a reasoning enhancement phase to improve the logical coherence of the final justifications.
    }
    % \vspace{-0.025in}
    \label{method}
\end{figure*}

\section{Methodology}

\subsection{Overview}

To achieve explainable hateful video detection, we propose a novel Information Augmentation and Reasoning Enhancement (IARE) framework. 
Specifically, the framework consists of two phases: an information augmentation phase, which enriches the rationale by incorporating sufficient contextual information from the video content, and a reasoning enhancement phase, which improves the logical coherence of the justifications by leveraging both correct and incorrect reasoning paths.
Together, these phases ensure a comprehensive understanding of video content while providing explainable decision rationales.
The overall framework of the proposed IARE is illustrated in Figure 2.
In the following sections, we elaborate on the IARE in detail.

\subsection{Information Augmentation}

We aim to develop explainable hateful video detection models that achieve accurate predictions while providing contextual rationales for their judgments. 
We employ the multimodal large language model as the backbone, leveraging its reasoning and generation capabilities to produce justifications. 
However, existing general-purpose models still exhibit limited understanding of hateful content, often overlooking subtle yet critical contextual elements \cite{DBLP:conf/www/LinLGMWY24}. 
This significantly restricts the evidential richness of the generated rationales.
To address this, we introduce a reasoning enhancement phase that incorporates a multimodal chain-of-thought approach, integrating relevant video contextual information to guide the reasoning process. 
Rich contextual cues from the proposed datasets, including video captions summarizing the main content and annotations of multimodal harmful elements, are integrated with raw video data (i.e., frame sequences and text information) during the process.
The detailed instruction template is shown as follows:

\begin{tcolorbox}[colback=gray!3, colframe=black, arc=2mm, boxrule=0.25mm, boxsep=0mm]
% \small
\textbf{\textit{Instruction}}:\\
You are an expert in hateful video detection. Determine whether the given video contains hate speech by integrating the provided background information. Your response must include a binary prediction (\texttt{hate} or \texttt{non-hate}) and a concise, evidence-based rationale.\\
\textbf{\textit{Input Context}}:\\
- Video Frames: [\textit{I}]\\
- Video Text (ASR and OCR): [\textit{T}]\\
- Video Caption: [\textit{C}]\\
- Potential Textual Harmful Elements: [$H_t$]\\
- Potential Visual Harmful Elements: [$H_v$]\\
\textbf{\textit{Response Format}}:\\
Prediction: [hate/non-hate]\\
Rationale: [Your reasoning here]
\end{tcolorbox}

To fully integrate contextual information with raw video content, we employ supervised instruction tuning to train the model, ensuring a comprehensive understanding of video semantics. 
To reduce computational costs and improve training efficiency, we introduce LoRA (Low-Rank Adaptation) training techniques.
The loss function for this process is defined as follows:

\begin{equation}
\small
\mathcal{L}_{\text{SFT}}
= - \mathbb{E}_{(X, Y)} 
\left[ \log \pi_{\theta}(Y \mid X) \right],
\end{equation}where $X$ and $Y$ denote the overall input and the gold rationale, respectively, and $\pi_{\theta}$ denotes the model being optimized. This training process encourages the model to more effectively leverage contextual information during decision-making, improving both detection accuracy and the evidentiary richness of its generated rationales.

\subsection{Reasoning Enhancement}

Although the training during the information augmentation phase can effectively enhance a model's understanding of video multimodal semantics, the model may still acquire potential biases from the training data, especially the spurious correlation between certain contextual information and labels. 
This "shortcut learning" mechanism causes the model to overly rely on superficial features rather than deep causal logic, thereby weakening its ability to distinguish boundary cases between hate and non-hate content and limiting the model's explainability.
To address this issue, we further introduce a reasoning enhancement stage into the training process. 
This method explicitly compares the differences between correct and incorrect reasoning paths, guiding the model to identify and discard spurious associations, thereby aligning it with correct reasoning patterns. Below is a detailed introduction.

We first construct preference pairs consisting of both correct and incorrect reasoning paths. 
The correct paths are taken directly from the gold rationales. 
To obtain incorrect paths, we prompt Qwen2.5-VL (72B) to generate explanations that support intentionally misassigned labels. 
By compelling the model to identify and combine all plausible video cues that could justify the incorrect label, the resulting explanations inherently capture potential spurious correlations.
We further conduct a manual inspection to ensure that these paths indeed constitute incorrect reasoning and use them as negative candidates in the preference pairs.
Notably, for unambiguous samples, where the video evidence strongly supports the gold label, the model cannot produce a coherent explanation for the intentionally incorrect label. We therefore filter out such samples at this stage. 
This mechanism automatically selects challenging cases containing easily confusable elements, thereby significantly enhancing the overall quality of the constructed sample pairs. As a result, the model can more effectively learn discriminative features from these samples.

On this basis, we utilize the sample pairs to further post-train the model that has passed through the information augmentation phase.
Specifically, we adopt the Direct Preference Optimization (DPO) method, which directly optimizes the model's reasoning capability by contrasting differences between various reasoning paths. 
During implementation, we use the same instruction template $X$ as in the information augmentation stage, which requires the model to output judgment results and decision rationales, and pair it with correct and incorrect reasoning paths, denoted as $Y^+$ and $Y^-$, respectively. 
We also employ LoRA technology for training.
The DPO loss function is defined as follows:

\begin{equation}
\small
\mathcal{L}_{\text{DPO}} = - \mathbb{E}_{(X, Y^+, Y^-)} \left[ 
    \log \sigma \left( 
        \beta \log \frac{\pi_{\theta}(Y^+ \mid X)}{\pi_{\text{ref}}(Y^+ \mid X)}
        - \beta \log \frac{\pi_{\theta}(Y^- \mid X)}{\pi_{\text{ref}}(Y^- \mid X)} 
    \right) 
\right],
\end{equation} where $\pi_{\theta}$ is initialized as the model that has undergone the information augmentation phase, $\pi_{\text{ref}}$ represents the reference model, a frozen-parameter version of $\pi_{\theta}$, and $\beta$ is a temperature parameter controlling the deviation from the reference policy. 
To improve training stability, we incorporate the negative log-likelihood loss between the model’s responses and the correct reasoning paths as an auxiliary regularization term following \cite{DBLP:conf/nips/PangYHCSW24}, computed as in Eq. (1). 
The final loss function during the knowledge discrimination phase is defined as follows:

\begin{small}
\begin{equation}
\mathcal{L}_{\text{KD}} = \mathcal{L}_{\text{DPO}} + \mathcal{L}_{\text{SFT}}.
\end{equation}
\end{small}This optimization objective reinforces discriminative features between different reasoning paths, encouraging the model to increase the generation probability of correct reasoning while suppressing incorrect reasoning, thereby effectively reducing the model's reliance on spurious associations of certain contextual cues.

\section{Experimental Setup}

\subsection{Dataset Splitting and Metrics}
\textbf{Dataset Splitting.} 
We validate the performance of the proposed IARE framework on both Ex-HateMM and Ex-ImpliHateVid. 
To ensure a fair comparison with the baselines, we adopted the same dataset split as the original datasets: HateMM \cite{DBLP:conf/icwsm/DasRSMG023}\footnote{https://github.com/hate-alert/HateMM} and ImpliHateVid \cite{DBLP:conf/acl/RehmanBKBK25}\footnote{https://github.com/videohatespeech/Implicit\_Video\_Hate}. 
For the Ex-HateMM dataset, the data is randomly split into training, validation, and test sets with a 7:1:2 ratio.
For the Ex-ImpliHateVid dataset, we directly adopted the official split provided in the original study.
Table \ref{dataset} summarizes the subset partitions of these two datasets.

\textbf{Metrics.} 
We use accuracy (Acc.), precision (P), recall (R), and macro F1-score (F1) as evaluation metrics, which are commonly used for hateful video detection tasks. 
Specifically, we select macro F1 score as the main metric, since it provides a comprehensive evaluation of both hateful and non-hateful samples, regardless of label imbalance in the dataset (see Table \ref{dataset}).

\begin{table}
  \centering
    \caption{Dataset splitting of hateful video datasets.}\label{dataset}
      % \vspace{-0.05in}
  \begin{tabular}{c|cc|cc}
    \toprule
    \textbf{Split} & \multicolumn{2}{c|}{\textbf{HateMM}} & \multicolumn{2}{c}{\textbf{ImpliHateVid}} \\
    & \textbf{\#Hate} & \textbf{\#Non-hate} & \textbf{\#Hate} & \textbf{\#Non-hate} \\
    \midrule
    Train & 293 & 456 & 607 & 598 \\
    Dev & 42 & 65 & 200 & 200 \\
    Test & 84 & 130 & 200 & 200 \\
    % \hline
    % \hline
    % Total & 1,064 & 1,949 & 124 & 230 \\
    \bottomrule
  \end{tabular}
  % \vspace{-0.05in}
\end{table}

% \begin{table}
%   \centering
%   \caption{Ablation studies by removing components from our proposed M2KE framework. Results show the macro F1 score.}
%     \begin{tabular}{m{1.5cm}|m{1.5cm}<{\centering}|m{1.5cm}<{\centering}|m{1.5cm}<{\centering}}
%     \toprule
%     Dataset & \multicolumn{1}{c|}{FHM} & \multicolumn{1}{c|}{MAMI} & \multicolumn{1}{c}{HarM} \\
%     % \midrule
%     % Model & \multicolumn{1}{c|}{F1} & \multicolumn{1}{c|}{F1} & \multicolumn{1}{c}{F1} \\
%     \midrule
%     M2KE  & \multicolumn{1}{c|}{75.6} &       &  \\
%     \midrule
%      \hspace{0.25cm} w/o EK &       &       &  \\
%      \hspace{0.25cm} w/o MD &       &       &  \\
%      \hspace{0.25cm} w/o CI &       &       &  \\
%      \hspace{0.25cm} w/o IER &       &       &  \\
%     \bottomrule
%     \end{tabular}%
%   \label{ablative}%
% \end{table}%

\subsection{Baselines}

We compare our proposed IARE framework against multiple hateful video detection models to demonstrate its effectiveness.
This includes comparisons with universal unimodal pre-trained models, current competitive detection models, and LLM-based methods.
Below are the introductions of these baselines:

\textbf{Universal pre-trained baselines:}
\vspace{-0.02in}
\begin{itemize}
\item \textbf{BERT}: BERT \cite{DBLP:conf/naacl/DevlinCLT19} is used as the unimodal method to leverage the textual modality;
\item \textbf{ViT}: an unimodal visual-only model that processes video frames using Vision Transformer \cite{DBLP:conf/iclr/DosovitskiyB0WZ21}; 
\item \textbf{MFCC}: a detection method that uses only audio signals with Mel-Frequency Cepstral Coefficients \cite{DBLP:conf/eccv/OwensE18};
\end{itemize}

\textbf{Competitive detection models:}
\vspace{-0.02in}
\begin{itemize}
\item \textbf{Concat}: concatenate features extracted by BERT, ViT, and MFCC and feed them into a classifier  \cite{DBLP:conf/icwsm/DasRSMG023};
\item \textbf{MulT}: a multimodal detection method that makes decisions by retrieving similar videos \cite{DBLP:conf/acl/TsaiBLKMS19};
\item \textbf{CSID}: a multimodal method that integrates self-supervised and supervised contrastive learning for detection \cite{DBLP:journals/kbs/LiLZLCSJ24}.
\item \textbf{MoRE}: a multimodal detection method that makes decisions by retrieving similar videos \cite{DBLP:conf/www/LangHXLX025};
\item \textbf{TCL}: a multimodal method that integrates self-supervised and supervised contrastive learning for detection \cite{DBLP:conf/acl/RehmanBKBK25}.
\end{itemize}

\textbf{LLM-based methods:}
\vspace{-0.02in}
\begin{itemize}
\item \textbf{GLM4.1V}: a multimodal LLM developed by Zhipu AI, and we adopt the GLM-4.1V-Thinking (9B) in the experiments;
\item \textbf{Qwen2.5-VL}: a multimodal LLM proposed by Alibaba, and we use it as another strong backbone.
\end{itemize}

We integrate IARE into both LLMs. In our main experiments, we also evaluate the LLMs under both zero-shot inference and supervised instruction fine-tuning settings for comparison, where the models directly output binary classification results.

% and few-shot settings (with two randomly selected examples, one hateful and one non-hateful),

\subsection{Implementation Details}

We train the model on the training set, evaluate its performance on the validation set, select the best-performing model parameters based on the validation set results, and then test the model on the test set.
To ensure real-world applicability, we rely on automatically derived auxiliary signals during inference for the validation and test sets. 
Specifically, GPT-4o-mini is used to generate video captions and extract harmful elements instead of using gold annotations. 
Validations on the test set show that 84\% of the generated information aligns well with our ground-truth annotations.

In the information augmentation phase, during SFT, the learning rate is set to 1e-4, and training is conducted for 2 epochs. 
In the reasoning enhancement phase, during DPO, the learning rate is set to 5e-6, with a beta value of 0.3, and training is conducted for 1 epoch. 
% Within the training sets of HateMM and ImpliHateVid, Qwen2.5-VL (72B) generates 627 and 458 incorrect reasoning paths, respectively, which are used as negative candidates in the preference pairs.
When using LoRA, the LoRA rank is set to 8, the LoRA alpha is set to 16, and the dropout rate is set to 0.05.
Additionally, the length of the generated rationale is constrained to be within 100 tokens.
Four NVIDIA H20 GPUs are used for training and inference.

% For the HateMM split, the training runtimes for SFT and DPO are 35 minutes and 20.8 minutes, respectively. For ImpliHateVid, the corresponding runtimes are 55 minutes and 14.9 minutes.

\begin{table*}[htbp]
  \centering
  \caption{Hateful video detection results on the two datasets. Results showing accuracy (Acc.), precision (P), recall (R), and macro F1 score (F1) (\%), where the \textbf{bold} and \underline{underline} scores respectively represent the optimal and suboptimal values. The baseline model results are reported by \cite{DBLP:conf/icwsm/DasRSMG023, DBLP:conf/acl/RehmanBKBK25, DBLP:conf/www/LangHXLX025}. All results are statistically significant, as determined by a $t$-test ($p < 0.01$).}
    \begin{tabular}{m{2.25cm}<{\centering}|m{2.75cm}<{\centering}|m{1cm}<{\centering}m{1cm}<{\centering}m{1cm}<{\centering}m{1cm}<{\centering}|m{1cm}<{\centering}m{1cm}<{\centering}m{1cm}<{\centering}m{1cm}<{\centering}}
    \toprule
          & \textbf{Dataset} & \multicolumn{4}{c|}{\textbf{Ex-HateMM}}   & \multicolumn{4}{c}{\textbf{Ex-ImpliHateVid}} \\
    \midrule
    Modality & Model & Acc.  & P     & R     & F1    & Acc.  & P     & R     & F1 \\
    \midrule
    \multirow{3}[2]{*}{Unimodal} & BERT  & 73.50  & 67.50  & 66.70  & 72.20  & 69.07  & 69.07  & 69.07  & 68.84  \\
          & ViT   & 74.80  & 69.50  & 65.60  & 67.20  & 76.55  & 76.58  & 76.56  & 76.84  \\
          & MFCC  & 67.50  & 59.30  & 67.90  & 66.50  & 49.87  & 24.93  & 50.00  & 66.55  \\
    \midrule
    \multirow{5}[2]{*}{Multimodal} & TTM   & 79.80  & 74.20  & 75.80  & 79.00  & 80.58     & 78.45     & 80.62     & 80.53 \\
          & MulT  & 65.71  & 43.18  & 65.71  & 52.12  & 83.52  & 83.20  & 83.80  & 83.52  \\
          & CSID  & 73.20  & 72.00  & 72.30  & 71.40  & 81.50  & 80.82  & 82.33  & 81.54  \\
          & MoRE  & 83.41  & 81.78  & 82.35  & 82.35  & -     & -     & -     & - \\
          & TCL   & 85.32  & 78.95  & 86.21  & 84.91  & 87.53  & 87.96  & 87.52  & 87.73  \\
    \midrule
    \multirow{6}[2]{*}{LLM-based} & GLM4.1V (9B) & 74.31  & 83.67  & 46.07  & 70.31  & 56.50  & 90.63  & 14.50  & 47.18  \\
          % & w/ few-shot & 84.40  & 82.72  & 77.01  & 83.54  & 86.25  & 87.96  & 85.00  & 85.93  \\
          & w/ SFT & 84.40  & 82.72  & 77.01  & 83.54  & 86.25  & 87.96  & 85.00  & 85.93  \\
          & w/ IARE & \underline{87.61}  & 84.88  & 83.91  & \underline{87.06}  & 88.00  & 90.86  & 84.50  & 87.99  \\
          & Qwen2.5-VL (7B) & 77.13  & 65.74  & \underline{88.32}  & 77.01  & 77.50  & \textbf{95.08 } & 58.00  & 76.61  \\
          % & w/ few-shot & 84.40  & 82.72  & 77.01  & 83.54  & 86.25  & 87.96  & 85.00  & 85.93  \\
          & w/ SFT & 86.70  & \textbf{87.18 } & 78.16  & 85.86  & \underline{89.50}  & 90.72  & \underline{88.00}  & \underline{89.50}  \\
          & w/ IARE & \textbf{90.37 } & \underline{85.42}  & \textbf{92.13} & \textbf{90.14 } & \textbf{91.75} & \underline{91.13}  & \textbf{92.50} & \textbf{91.75} \\
    \bottomrule
    \end{tabular}%
  \label{baseline}%
\end{table*}%

\section{Results and Analysis}

\subsection{Overall Performance}

Table \ref{baseline} demonstrates the performance of our proposed IARE framework with the compared hateful video detection methods on the two datasets, HateMM and ImplicHateVid. 
Based on the results, we can draw the following conclusions:

(1) Overall, when considering the modality of input information, the multimodal detection methods presented in the second block of Table \ref{baseline} clearly outperform unimodal models, including BERT, ViT, and MFCC. This highlights the importance of integrating elements from different modalities for effective detection of hateful videos.
Moreover, multimodal LLMs demonstrate strong performance, even in zero-shot scenarios, where their detection accuracy on two datasets is comparable to that of baseline models specifically trained for certain tasks, such as TTM and CSID. This underscores the model's capacity to comprehend video content. Following the introduction of instruction-based fine-tuning, the model’s performance is further enhanced, providing a solid foundation for our research on explainable hate video detection.

(2) Compared to other competitive task-specific baselines, our proposed IARE framework demonstrates a notable improvement in detection performance, consistently achieving state-of-the-art (SOTA) results on the HateMM and ImpliHateVid datasets.
Taking the experimental results on Qwen-2.5 (7B) as an example, compared to the backbone that has undergone supervised fine-tuning, the introduction of IARE leads to increases in macro F1 scores of 4.28\% and 2.25\%, respectively.
In comparison to MoRE, which uses multimodal features through neural network modules with limited explainability, and TCL, which combines coarse-grained sentiment information from the video, IARE more effectively enhances the detection of hateful videos. 
This improvement is attributed to IARE's explainable detection mechanism, which facilitates more effective reasoning by outputting the rationale behind its decisions, rather than simply learning the direct mapping between samples and labels.
Additionally, this underscores the importance of integrating rich contextual information and ensuring reasoning coherence, which contributes to a more comprehensive understanding of hateful videos.
Next, we will conduct a detailed analysis of each module in IARE through ablation studies.

\subsection{Ablative Studies}

We conduct ablative studies on several variants of IARE:
1) \textit{w/o Information Augmentation (IA)}: Directly introduce the reasoning enhancement phase to train the original backbones and obtain the prediction results;
2) \textit{w/o Multimodel Chain-of-Thought (MCoT)}: The model is instructed to output only the final result (i.e., hateful or non-hateful) without providing the underlying rationale;
3) \textit{w/o Video Caption (VC)}: The multimodel Chain-of-Thought excludes the video caption;
4) \textit{w/o Harmful Elements (HE)}: The instruction template excludes any potentially harmful elements;
5) \textit{w/o Reasoning Enhancement (RE)}: Directly use the model trained through the information augmentation phase to infer;
6) \textit{w/o Incorrect Reasoning Path (IRP)}: Replace incorrect reasoning paths with judgments that only contain incorrect labels, and pair them with correct reasoning paths to form sample pairs for DPO training.
Ablation studies were conducted using Qwen2.5-VL, which exhibits strong detection performance.
The ablation results are presented in Figure 3.

\begin{figure}
    \centering
    \includegraphics[width=8.25cm]{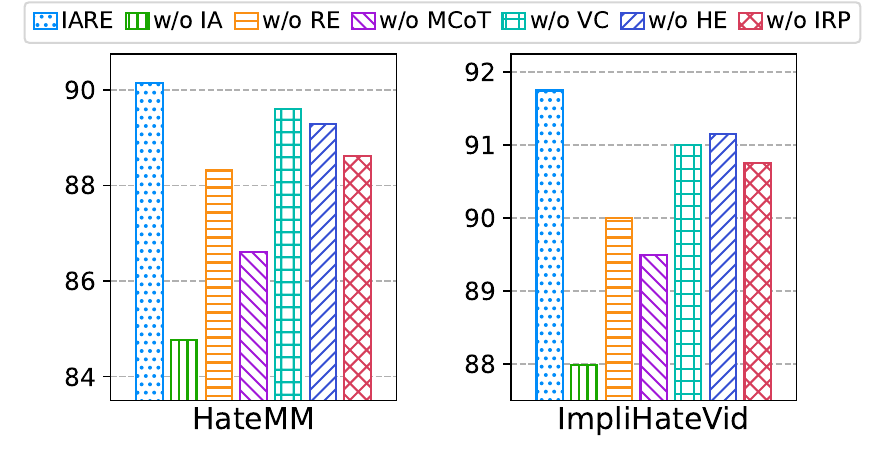}
    % \vspace{-0.15in}
  \caption{Ablation studies by removing components from our IARE framework. Results show the macro F1 score.}
    % \vspace{-0.2in}
    \label{Ablation}
\end{figure}

Based on the results, we note that the ablative models suffer varying degrees of performance degradation.
This indicates the effectiveness of our proposed components for hateful video detection.
Specifically, in the comparison of the two main stages, the information augmentation phase (\textit{IA}) brings a clearer performance improvement compared to the reasoning enhancement phase (\textit{RE}), with relative improvements of 5.03\% and 2.00\%, respectively. 
This is because the primary goal of the information augmentation phase is to help the model master the foundational skills required for hateful video detection. 
The training in this phase lays the groundwork for the subsequent reasoning enhancement phase, making it a more crucial factor in the overall improvement.
Among the submodules, the multimodal chain-of-thought (\textit{MCoT}) provides the most improvement (a relative increase of 3.21\%), indicating its effective integration of multimodal reasoning to enhance the model’s understanding of video content. 
It also highlights that requiring the model to output explanations helps it incorporate the full context, improving its performance in detection.
On the other hand, incorporating video captions (\textit{VC}) and harmful elements (\textit{HE}) resulted in the smallest improvements (a relative increase of 0.72\% and 0.80\%, respectively), likely because the multimodal LLM already possesses strong video analysis capabilities. 
However, despite the limited performance gain, adding this information still enhances the explainability of the IARE framework's decision-making process, which will be further discussed in the following section.

\subsection{Effect of LLM Size}

In this section, we analyze the performance of IARE for explainable hateful video detection across LLMs of different sizes. 
We adopt Qwen2.5-VL as the backbone, given its strong performance in the main experiments. 
We evaluate three variants (3B, 7B, and 32B). 
In addition to IARE, we report results under zero-shot inference, SFT, and DPO to provide a comprehensive comparison. 
The results are shown in Table~\ref{size}.

Based on the results, we can observe that: 
1) Under the zero-shot setting, model scale has a substantial impact on hateful video detection performance. Compared with Qwen2.5-VL (7B), the 3B variant underperforms by an average of 13.0\% in macro-F1 across the two datasets, while the 7B model further lags behind the 32B variant by 5.9\%. This trend reveals a clear scaling effect: larger models exhibit stronger multimodal video understanding, particularly in zero-shot inference, where performance relies heavily on the model’s intrinsic reasoning capability.
2) After instruction fine-tuning, the three model sizes achieve broadly comparable detection performance on both datasets, with only a marginal gap between the 32B and 3B variants. This is mainly because the objective is reduced to predicting rigid binary labels, which directly supervises the final decision and thus lessens the need for complex multimodal reasoning. As a result, the decision boundary becomes easier to learn, enabling even the 3B model to perform close to larger counterparts after supervision.
3) By contrast, applying DPO alone yields only moderate improvements, because preference optimization mainly focuses on challenging samples. Without prior supervised grounding, training directly on these instances provides high-variance signals, making it difficult for the model to acquire generalizable multimodal patterns of hateful videos. Instead, it tends to overfit to local decision boundaries, resulting in limited gains.
4) After integrating the full IARE, the scaling effect becomes evident again, with the 7B and 32B models consistently outperforming the 3B variant. This is because IARE shifts the task from simple label prediction to evidence-grounded, coherence-driven reasoning, requiring multimodal integration and inference that benefit from larger model scale. Consequently, larger models can better leverage the structured evidence and reasoning signals introduced by IARE, yielding more pronounced gains than standard binary instruction fine-tuning.

\begin{table}
  \centering
  \caption{Effect of LLM size under different settings. All experiments adopt Qwen2.5-VL as the backbone, and performance is evaluated using macro F1.}
  % \vspace{-0.05in}
    \begin{tabular}{c|m{0.8cm}<{\centering}m{0.8cm}<{\centering}m{0.8cm}<{\centering}|m{0.8cm}<{\centering}m{0.8cm}<{\centering}m{0.8cm}<{\centering}}
    \toprule
    Dataset & \multicolumn{3}{c|}{Ex-HateMM} & \multicolumn{3}{c}{Ex-ImpliHateVid} \\
    \midrule
    Model & 3B    & 7B    & 32B   & 3B    & 7B    & 32B \\
    \midrule
    Base  & 55.50  & 77.01  & 82.10  & 72.09  & 76.61  & 83.26  \\
    w/ SFT  & 85.94  & 85.86  & 86.14  & 88.69 & 89.50  & 89.99  \\
    w/ DPO  & 54.72 & 77.45 &  82.30  & 71.48 & 77.50  &  84.50 \\
    IARE  & 87.57  & 90.14  & 90.47  & 89.24 & 90.75  & 92.25  \\
    \bottomrule
    \end{tabular}%
  % \vspace{-0.05in}
  \label{size}%
\end{table}%

\subsection{Evaluation of Explainability}

In this section, we evaluate the quality of the rationales generated by the IARE framework, further demonstrating its improvement in explainability. 
Since traditional evaluation metrics, such as ROUGE and BERT-Score, are unsuitable for assessing the reliability of rationales with non-unique gold labels, we adopt the evaluation framework proposed by \cite{DBLP:conf/ijcai/WangHACL23}, analyzing the rationales across the following dimensions:
1) \textit{Informativeness}: The rationale provides new information, such as explaining the background and additional context;
2) \textit{Readability}: The rationale adheres to proper grammar and structural rules;
3) \textit{Soundness}: The rationale is logically valid and well-supported;
4) \textit{Persuasiveness}: The rationale is clear and convincing.
For each dimension, a 5-point Likert scale was employed, where 1 indicates the poorest quality and 5 represents the best.
In our implementation, we evaluate the rationales generated by Qwen2.5-VL (7B) on a subset of 200 samples randomly selected from the test sets of Ex-HateMM and Ex-ImpliHateVid. 
We compare three settings: the original backbone, a variant integrating the information augmentation phase (\textit{w/ IA}), and our proposed IARE framework.
To ensure reliable assessment, both automatic and manual evaluation adopted. 
For automatic evaluation, we use GPT-4o as the evaluator, which has shown strong reliability in rationale assessment \cite{DBLP:conf/www/LinLGMWY24}. 
For manual evaluation, two annotators independently score each rationale, and the averaged score is reported. 
The ground-truth rationales are also used as reference during evaluation.
The results are shown in Table \ref{evaluation}.

Based on the results, we observe that: 
1) Across various metrics, the rationales generated by IARE demonstrate consistent improvements over the initial ones generated by the backbone, highlighting their enhanced quality and contribution to the model's understanding of videos.
2) Specifically, after integrating the augmented information, we observe a notable gain in \textit{Informativeness}, with average relative improvements of 11.6\% in both automatic and human evaluations, highlighting the importance of evidence-guided reasoning. 
3) Additionally, \textit{Persuasiveness} further increases after introducing the reasoning enhancement stage, yielding a 9.4\% improvement. This suggests that contrasting positive and negative rationale paths helps the model better identify key cues, rather than relying on spurious associations.
4) By contrast, the improvement in Readability is relatively modest, as LLMs already exhibit strong text generation abilities, producing outputs with few grammatical errors and inherently good readability.

% \begin{table}
%   \centering
%   \caption{Evaluation of the rationale quality on hateful videos in the test set, including Informativeness (Info.), Readability (Read.), Soundness (Sound.), and Persuasiveness (Pers.).}
%   \begin{tabular}{
%     >{\centering\arraybackslash}m{1.25cm}|
%     >{\centering\arraybackslash}m{0.75cm}
%     >{\centering\arraybackslash}m{0.75cm}
%     >{\centering\arraybackslash}m{0.75cm}|
%     >{\centering\arraybackslash}m{0.75cm}
%     >{\centering\arraybackslash}m{0.75cm}
%     >{\centering\arraybackslash}m{0.75cm}
%   }
%     \toprule
%     Evaluator & \multicolumn{3}{c|}{GPT-4o} & \multicolumn{3}{c}{Human} \\
%     \midrule
%     Rationales & Qwen & w/~IA & IARE & Qwen & w/~IA & IARE \\
%     \midrule
%     Info. & 2.86 & 3.97 & 4.05 & 3.26 & 4.28 & 4.32 \\
%     Read.     & 4.25 & 4.28 & 4.33 & 4.60 & 4.62 & 4.63 \\
%     Sound.    & 3.24 & 3.60 & 3.81 & 3.54 & 3.86 & 3.95 \\
%     Pers.  & 3.07 & 3.74 & 4.13 & 4.23 & 3.26 & 4.13 \\
%     \bottomrule
%   \end{tabular}
%   \label{evaluation}
% \end{table}

\begin{table}
  \centering
  \caption{Evaluation of the rationale quality on hateful videos in the test set, including Informativeness (Info.), Readability (Read.), Soundness (Sound.), and Persuasiveness (Pers.).}
  % \vspace{-0.05in}
  \begin{tabular}{
    >{\centering\arraybackslash}m{1.25cm}|
    >{\centering\arraybackslash}m{0.75cm}
    >{\centering\arraybackslash}m{0.75cm}
    >{\centering\arraybackslash}m{0.75cm}|
    >{\centering\arraybackslash}m{0.75cm}
    >{\centering\arraybackslash}m{0.75cm}
    >{\centering\arraybackslash}m{0.75cm}
  }
    \toprule
    Evaluator & \multicolumn{3}{|c|}{GPT-4o} & \multicolumn{3}{c}{Human} \\    \midrule
    Rationales & Qwen  & w/ IA & IARE  & Qwen  & w/ IA & IARE \\
    \midrule
    Info. & 3.43  & 3.85  & 3.88  & 3.59  & 3.98  & 4.05 \\
    Read. & 4.15  & 4.18  & 4.17  & 4.71  & 4.73  & 4.76 \\
    Sound. & 3.54  & 3.69  & 3.80  & 3.65  & 3.82  & 3.93 \\
    Pers. & 3.37  & 3.58  & 3.92  & 3.66  & 3.86  & 4.22 \\
    \bottomrule
    \end{tabular}%
  % \vspace{-0.05in}
  \label{evaluation}%
\end{table}%

\setcounter{figure}{5}
\captionsetup[figure]{name=Table}
\begin{figure*}[t]
    \centering
    \includegraphics[width=17.5cm]{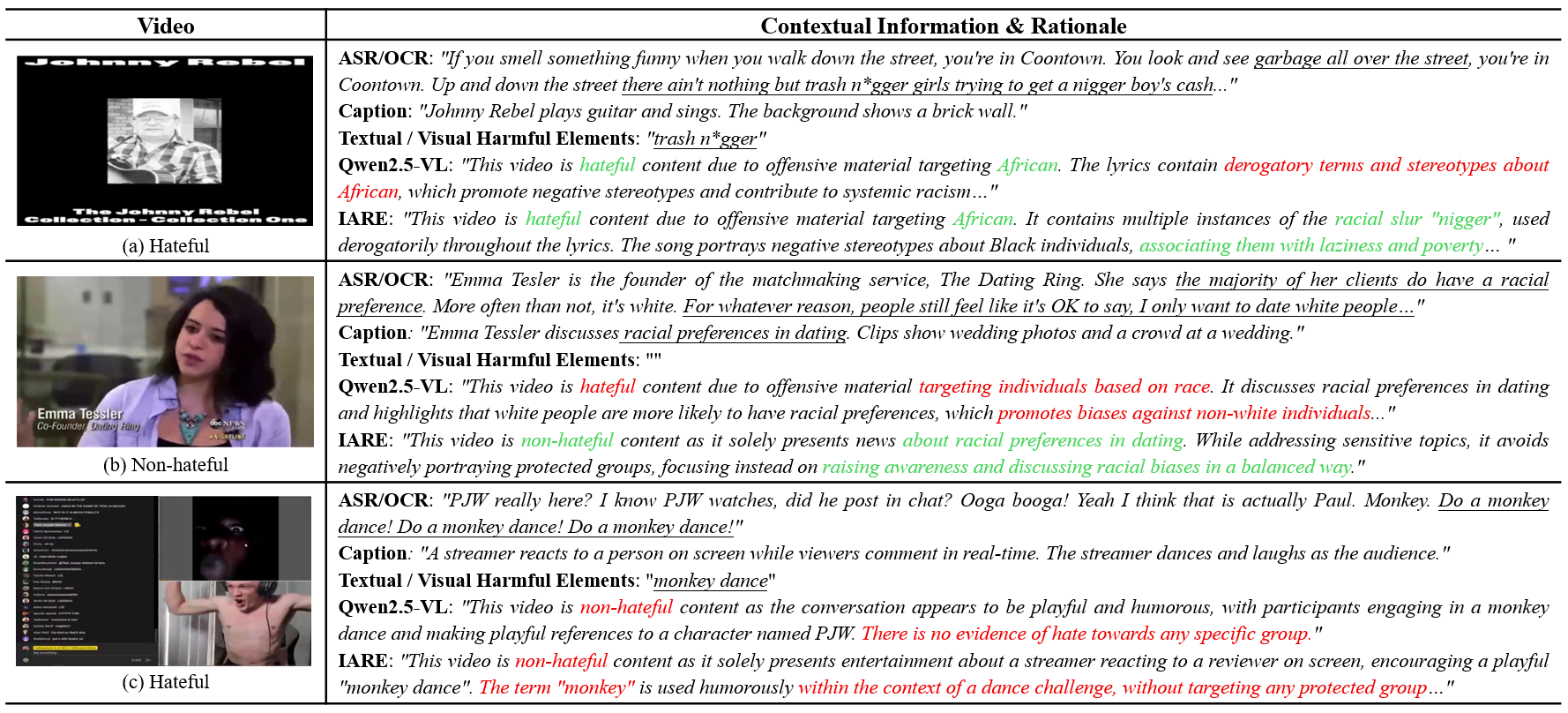}
    \vspace{-0.025in}
    \caption{Illustration of the case study. The specific content in the video that significantly impacts the decision-making process is highlighted with \underline{underline}. 
    Accurate predictions and supporting cues are highlighted in \textcolor{LimeGreen}{green}, while inaccuracies or statements using spurious evidence are marked in \textcolor{Red}{red}.}
    \vspace{-0.025in}
    \label{case_study}
\end{figure*}

\subsection{Case Study}

To further examine how our proposed IARE improves the explainability of detection models, we conduct two case studies, as shown in Table \ref{case_study}(a) and (b). Specifically, the two instances are randomly sampled from the two datasets. For each instance, we provide the dataset’s contextual information as reference and report the rationales generated by the original Qwen2.5-VL for comparison. These cases are used to highlight IARE’s typical advantages in rationale generation, including more specific evidence grounding and improved logical coherence, as detailed below.

\textbf{More specific evidence}: For Exp. (a), a hateful video, both Qwen2.5-VL and IARE provide the correct judgment.
Nevertheless, Qwen2.5-VL uses generalized descriptions when generating rationales, only stating that the video contains derogatory terms and stereotypes about Africans, without providing specific harmful content or details. 
While this might be due to safety considerations for the open-sourced model, it results in limited explainability, as clear rationales are essential for real-world applications.
In contrast, IARE incorporates the lyrics from the video, explicitly revealing the underlying metaphor that links “\textit{laziness and poverty}” to Africans, thereby enhancing the transparency of the model's decision-making process.
This demonstrates the necessity of the information augmentation phase, where introducing contextual information enriches the evidence supporting the rationale.

\textbf{Improved logical coherence}: For Exp. (b), a non-hateful video, Qwen2.5-VL mistakenly classifies the video as hateful. 
This occurs because the video addresses the sensitive topic of "\textit{racial preferences in dating}" among white people, despite presenting it objectively.
However, Qwen2.5-VL mistakenly interprets it as racial discrimination.
This highlights an issue of over-sensitivity in LLMs, which struggle to distinguish between benign discussions of sensitive social issues and offensive content, thereby affecting the model's detection accuracy.
In contrast, our proposed IARE accurately evaluates the video and provides a neutral rationale. 
This is due to the model's reasoning enhancement phase, where it utilizes the full context to make logically consistent inferences, reducing reliance on specific elements and topics. 
This approach improves the model's understanding and enhances detection accuracy.

In addition, we further discuss the limitations of IARE by analyzing samples it misclassifies. Exp. (c) presents a representative hateful video. In this sample, a streamer interacting with African viewers is prompted by other users to perform a “monkey dance”, leading the streamer to engage in humorous movements. 
This behavior is considered hateful because the term “monkey” has been used in certain cultural contexts as a derogatory metaphor targeting Black individuals, and has historically functioned as a symbolic form of racial discrimination.
However, although IARE identifies the presence of the term “monkey”, it fails to perform deeper reasoning that connects this metaphorical expression to the hate context. 
As a result, the model incorrectly interprets the video as purely entertaining rather than offensive. This case highlights the model’s limited capability in understanding cultural symbols and implicitly discriminatory meanings. 
Therefore, improving models’ ability to reason about subtle and group-targeted hateful expressions remains a challenge for explainable hateful video detection.

\section{Conclusions}
In this paper, we study explainable hateful video detection, aiming to equip models with the ability to provide contextual rationales alongside their predictions. We introduce Ex-HateMM and Ex-ImpliHateVid, two explainable hateful video datasets annotated with fine-grained labels, including potential harmful elements and contextual rationales grounded in video content.
Building on these resources, we propose a novel Information Augmentation and Reasoning Enhancement (IARE) framework for explainable detection. IARE first performs an information enhancement phase that integrates multimodal cues to enrich the evidence underlying rationales, and then applies a reasoning enhancement phase to improve the logical coherence of the resulting justifications.
Experiments on both datasets show that IARE achieves state-of-the-art performance, highlighting the importance of explainable hateful video detection.
Further ablation studies and qualitative analyses confirm its capability to generate accurate and meaningful rationales that support its detection decisions.

\section{Acknowledgment}
This research was supported in part by the National Natural Science Foundation of China (Nos. 625B2033, 62376051, and 62576073), the Ministry of Education Humanities and Social Science Project of China (No. 25YJCZH308), and the National Research Foundation, Prime Minister’s Office, Singapore, together with the Ministry of Digital Development and Information, Singapore, under the Online Trust and Safety (OTS) Research Programme (Award No. S24T2TS007). The examples provided in this paper do not represent the views of the authors. 
Any conclusions do not necessarily reflect the views of the National Natural Science Foundation of China, the National Research Foundation, Prime Minister’s Office, Singapore, or the Ministry of Digital Development and Information, Singapore.

\bibliographystyle{ACM-Reference-Format}
\balance
\bibliography{base, explain_hate, LLM, hate_video, hate_meme}

\end{document}